\documentclass[12pt]{article}
\usepackage[english]{babel}
\usepackage{amssymb, amsfonts}
\usepackage{amsmath}
\usepackage{graphics,subfigure}
\usepackage{epsfig}
\title { Multidimensional data classification with artificial neural networks}
\author{F. Barbarino, P. Boinee, A. De Angelis \\
{\small {\it Dipartimento di Fisica, Universit\`a di Udine and INFN Trieste,}}\\ {\small {\it via delle scienze, 208, I-33100 Udine, Italy}
}}

\begin{document}
%\renewcommand{\ttdefault}{pcr}
%\include{title}
%\pagestyle{headings}
%\pagenumbering{roman}
\maketitle

\begin{abstract}
\noindent
Multi-dimensional data classification is an important and challenging problem in many astro-particle experiments. Neural networks have proved to be versatile and robust in multi-dimensional data classification. In this article we shall study the classification of gamma from the hadrons for the MAGIC Experiment. Two neural networks have been used for the classification task. One is Multi-Layer Perceptron based on supervised learning and other is Self-Organising Map (SOM), which is based on unsupervised learning technique. The results have been shown and the possible ways of combining these networks have been proposed to yield better and faster classification results.
\end{abstract}

\section{Introduction}
Many high-energy gamma ray experiments have to deal with the problem of separating gammas from hadrons ~\cite{p1}. The experiments usually generate large data sets with many attributes in them. This multi-dimensional data classification problem offers a daunting challenge of extracting small number of interesting events (gammas) from an overwhelming sea of background (hadrons) . Many techniques are in active research for addressing this problem. The list includes classical statistical techniques to more sophisticated techniques like neural networks, classification trees and kernel functions. 

The class of neural networks provides an automated technique for the classification of the data set into given number of classes~\cite{kk}. It is in active research in both artificial intelligence and machine learning communities.  Several neural network models have been developed to address the classification problem. Usually, one makes the distinction between supervised and unsupervised classifiers: A  supervised classifier is used , when an analyst has some examples, for which the correct classification is known. This can be done, for example, in most problems related to particle physics at accelerators, where there is a generally good knowledge of detectors and of the underlying physics, and good simulations are available. Whereas in an unsupervised technique,  the events are partitioned into classes of similar elements, without using additional information. This is the case especially for fields operating in a discovery regime, as, e.g., astroparticle physics~\cite{ale}.

From a mathematical perspective, a neural network is simply a mapping from $R^n \rightarrow R^m$, where $R^n$ is the input data set dimension and $R^m$ is the output dimension of the neural network . The network is typically divided into various layers; each layer has a set of neurons also called as nodes or information units, connected together by the links. The artificial neural networks are able to classify data by learning to discriminate patterns in features (or parameters) associated with the data.
 The neural network learns from the data set when each data vector from the input set is subjected to it. The learning or information gain is stored in the links associated with the neurons.
   
 The output generated by the network depends on both the problem and network type. For the gamma/hadron separation problem the supervised network maps each input vector onto the [0,1] interval, whereas in unsupervised networks the nodes are adapted to the input vector in such a way that the output of the network represents the natural groups that exist in the data set. A visualization technique is used to view the groups discovered by the network. 

Section 2 describes the data sets used for the classification. Section 3 deals with the multilayer perceptron network and its classification results. Section 4 deals with Self-Organizing maps and its variant along with their classification results. Conclusions and future perspectives have been discussed in the section 5.

\section{Data set description}
The data sets are generated by a montecarlo program, CORSIKA ~\cite{cor}. 
They contain 12332 gammas, 7356 'on' events (mixture of gammas and hadrons), 
and 6688 hadron events. These events are stored in different files. The files contain event parameters in ASCII format, each line of 12 numbers being one event~\cite{boc}, with the parameters defined below,  
                                                                                                        
\begin{enumerate}
\item   fLength:    major axis of ellipse [mm]
\item	fWidth:     minor axis of ellipse [mm]
 \item fSize:      10-log of sum of content of all pixels 
 \item  fConc:      ratio of sum of two highest pixels over fSize  [ratio]
\item fConc1:     ratio of highest pixel over fSize  [ratio]
\item fAsym:      distance from highest pixel to centre, 
                       projected onto major axis [mm]
\item fM3Long     3rd root of third moment along major axis  [mm]
\item fM3Trans    3rd root of third moment along minor axis  [mm]
\item fAlpha:     angle of major axis with vector to origin [deg]
\item fDist:      distance from origin to centre of ellipse [mm]
\item fEner:      10-log of MC energy [in GeV]
\item fTheta:     MC zenith angle [rad] 
\end{enumerate}
The first 10 image parameters are derived from pixel analysis, and are 
used for classification.
%------------------------------------------MLP----------------------------------------------------------------------
\section{Multi-Layer Perceptron}

For this approach we used the ROOT Analysis Package (v. 4.00/02) and in particular the MultiLayer Perceptron class~\cite{kn:mlp}, which implements a generic layered network. Since this is a supervised network we took half of Gamma and OFF data to train the network and the remaining data to test it. The code of the ROOT package is very flexible and simple to use. It allowed us to create a network with a 10 nodes input layer, a hidden layer with the same number of nodes and an output layer with just a single neuron which should return "0" if the data represent hadrons or "1" if they're gammas. Weights are put randomly at the beginning of the training session and then adjusted from the following runs in order to minimize errors (back-propagation). Errors at cycle $i$ are defined as: $err_i = \frac{1}{2} \; o_i^2$
where $o_i$ is the error of the output node. 
Data to input and output nodes are transferred linearly, while for hidden layers they use a sigmoid (usually: $\sigma(x) = 1/(1 + \exp(-x))$). 

We have tested the same network using different learning methods proposed by the code authors, as for example the so called "Stochastic minimization", based on the Robbins-Monro stochastic approximation, but the default "Broyden, Fletcher, Goldfarb, Shanno" method has proved to be the quickest and with the better error approximation.

\begin{figure}[hct]
%\centering
\mbox{\subfigure[The error functions for training and test data took on 1000 runs.]{\epsfig{figure=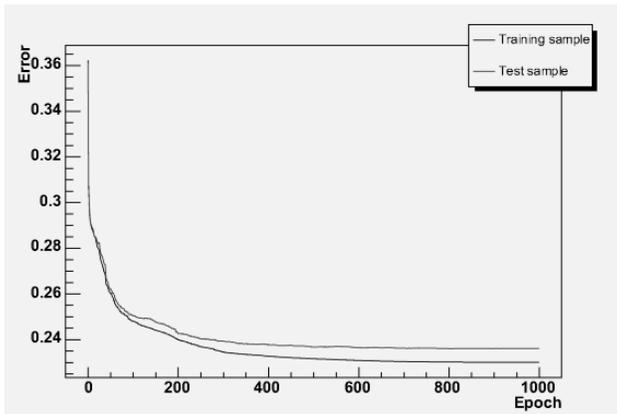,width=.60\textwidth}}\quad\subfigure[The histogram of distributions for gamma and hadron parameters.]{\epsfig{figure=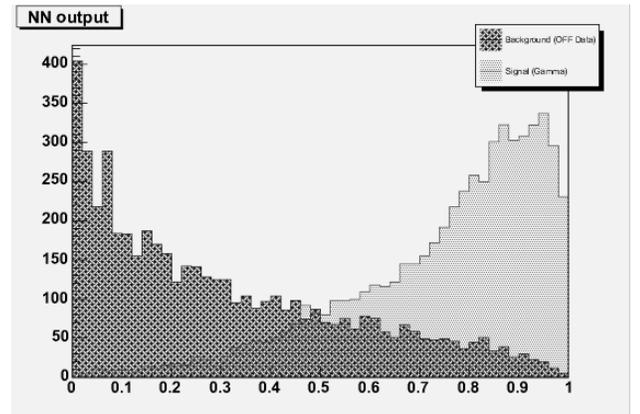,width=.60\textwidth}}}
\caption{MLP classification results} 
\end{figure}

Figures 1.a and 1.b  represent a possible output when using the ROOT package on those data. The first one depicts the error function for each run of the network, comparing the training and the test data. Note that the greater is the number of runs, the better the network behaves. 
The second one shows the distributions of output nodes, that is how many times the network decides to give a value near to "0" or to "1".

%--------------------------------------------------SOM--------------------------------------------------------------
\section {Self-Organising Maps (SOM)}

SOM is based on unsupervised learning technique. It is used in the classification of data sets with no labels. It consists of a map of information units also called as neurons, arranged in a two-dimensional grid~\cite{p2}. Every neuron $i$ of the map is associated with a $n$-dimensional reference vector $m_i = { [ { m_{i1},\ldots,m_{in} } ] }^T$ , where $n$ denotes the dimension of the input vectors. The neurons of the map are connected to adjacent neurons by a neighbourhood relation, which dictates the topology, or the structure, of the map. The most common topologies in use are rectangular and hexagonal.
The learning process of the SOM is as follows: 
\begin{enumerate}
\item {\bf Initialisation phase:} Initialise all the neurons in the map with the input vectors randomly.
\item {\bf Data normalization:} For a better identification of the groups the data have to be normalized. We employed the `range' method where each component of the data vector is normalized to lie in the intravel [0,1].
\item {\bf SOM Training:} Select an input vector $x$ from the data set randomly. A best matching unit (BMU) for this input vector, is found  in the map by the following metric 
	$$ 
	\left\| x- m_c \right\| = 
	\min_i \left\{ \left\| x- m_i \right\| \right\} 
	$$ 
where $m_i$ is the reference vector associated with the unit $i$.
\item {\bf Updating Step:}
The reference vectors  of BMU and its neighbourhood are updated according to the following rule
$$ 
m_i(t+1)=\left\{ 
\begin{alignedat}{2} 
&m_i(t) + \alpha(t)\cdot h_{ci}(t)\cdot[x(t)-m_i(t)], & & \qquad i \in N_c(t) 
\\ 
&m_i(t), & & \qquad i \notin N_c(t) 
\end{alignedat} 
\right. 
$$ 
where
\\ $h_{ci}(t) $ is the kernel neighbourhood around the winner unit $c$. \\
\\ $t$ is the  time constant.\\
\\$x(t)$ is an input vector randomly drawn from the input data set at time~$t$.\\
\\ $\alpha(t)$ is the learning rate at time~$t$. \\
%\\ $d^2_{ci}$ is the distance between the winner unit $c$ and the unit i in its neighbourhood.   \\
\\ $N_c(t)$ is the neighbourhood set for the winner unit $c$.\\
The above equation make BMU and its neighbourhood  move closer to the input vector. This adaptation to input vector forms the basis for the group formation in the map. 
\item {\bf Data groups visualisation:} steps 3 and 4 are repeated for selected number of trials or epochs. After the trails are completed the map unfolds itself to the distribution of the data set finding the number of natural groups exist in the data set. The output of the SOM is the set of reference vectors associated with the map units. This set is termed as a codebook.   To view the groups and the outliers discovered by the SOM we have to visualize the codebook. U-Matrix is the technique typically used for this purpose. 

\end{enumerate}

\noindent
 The ON events data set has directly used with the SOM. No prior training is required. The unsupervised behavior of SOM had discovered the groups in the data set in an automatic way. We worked with two  kernel neighbourhoods of the SOM that are described below. 
\begin{enumerate}
\item[\bf a)] {\bf Gaussian SOM }
\\The kernel neighbourhood  is defined by gaussian function\\
$h_{ci}(t) =e^{\LARGE{\left({d^2_{ci}}/{2\sigma^2_t}\right)}}, $ here $d^2_{ci}$ is the distance between the winner unit $c$ and the unit $i$, $\sigma_t$ is the neighbourhood radius.  
The results of the classification are shown in the figure~2.a. The Map is a 25X25 network and is trained with 300 epochs. Further increase in map size and epochs does not shown any improved results.
 
\item[\bf b)]  {\bf Cutgaussian SOM }
\\ The kernel neighbourhood is defined by cut-gaussian function\\ 
$h_{ci}(t) = e^{\LARGE{\left({d^2_{ci}}/{2\sigma^2_t}\right)}}\cdot1\left(\sigma_t-d_{ci}\right). $
The results of the classification are shown in the figure~2.b. The Map is 40X30 network and is trained with 300 epochs. Further increase in map size and epochs does not shown improved results.The cutgaussian kernel shown better performance than that of of gaussian kernel.

\end{enumerate}
\begin{figure}[hct]
%\centering
\mbox{\subfigure[SOM with gaussian kernel: Two groups are discovered that are seperated by outliers and boundaries.]{\epsfig{figure=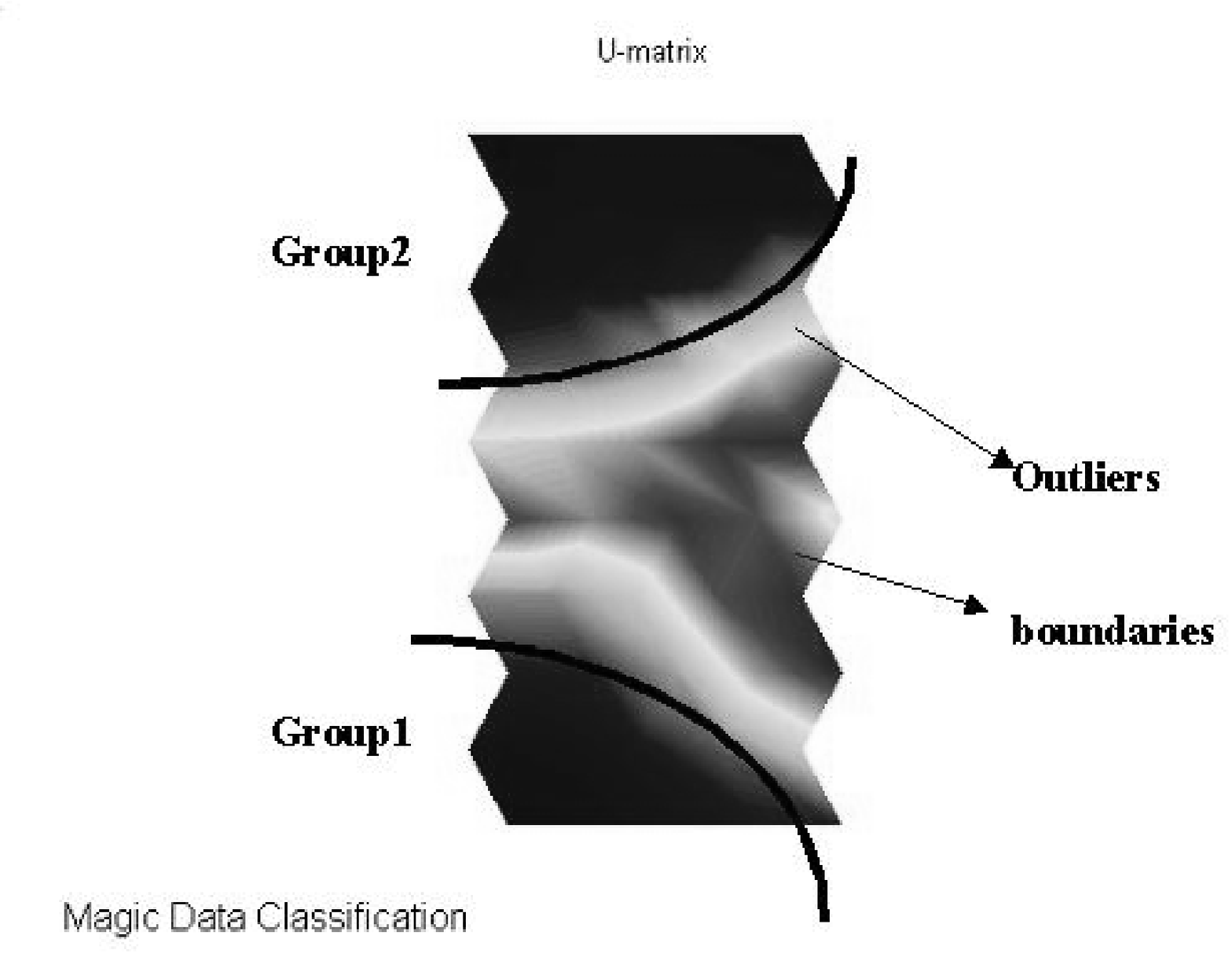,width=.60\textwidth}}\quad\subfigure[SOM with cutgaussian kernel: Again the algorithm found 2 groups and the outliers are well seperated this time.]{\epsfig{figure=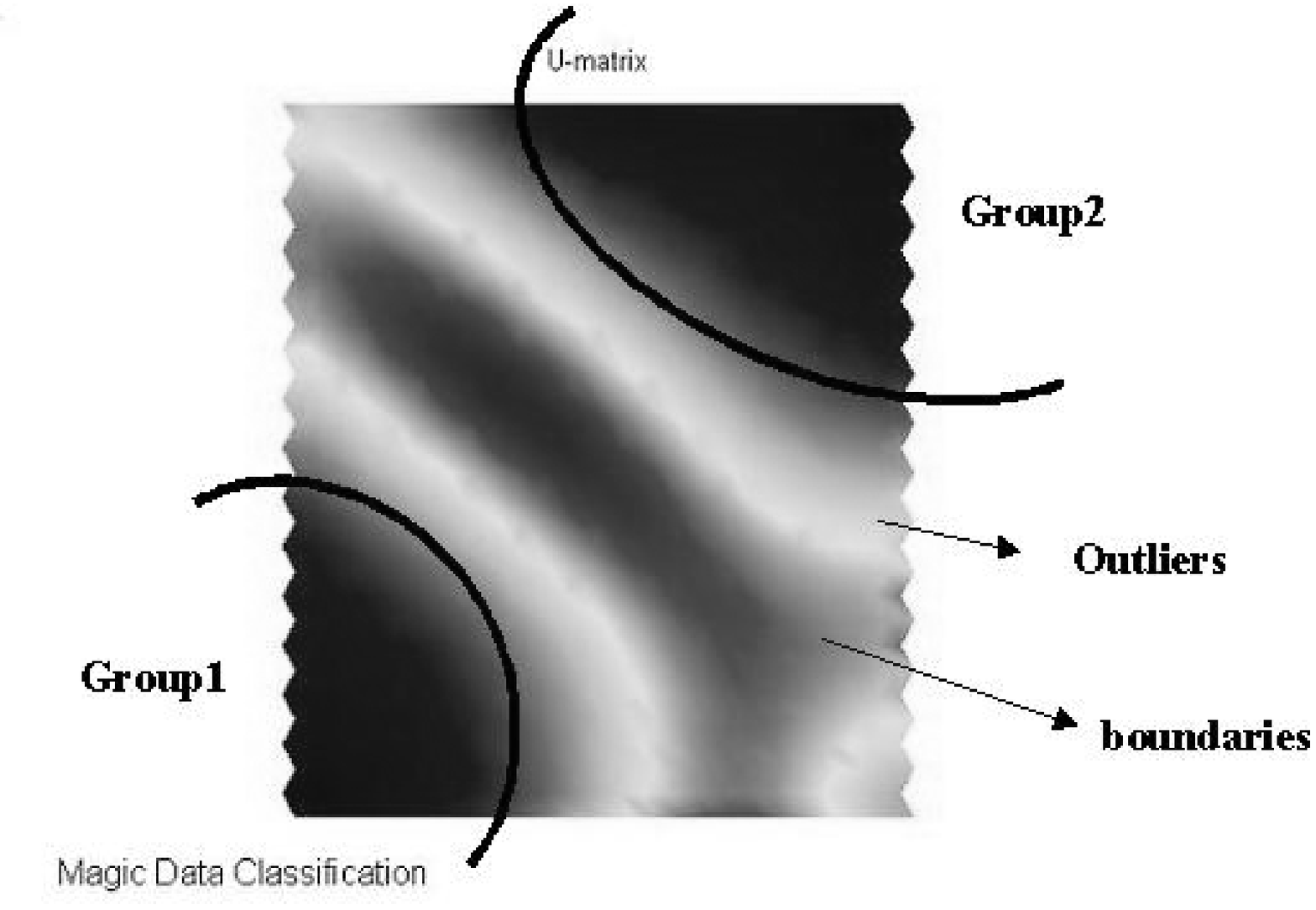,width=.60\textwidth}}}
\caption{SOM classification results} 
\end{figure}  
\noindent
We developed a c++ implementation of SOM with both kernel neighbourhoods. The SOM trained results are visualized using the u-matrix technique implemented in SOM TOOLBOX 2.0 in MATLAB environment~\cite{kn:vesa}. 
%\($$h_{ci}(t) = \exp^\frac{d^2_{ci}}{2\sigma^2_t}\cdot1\left(\sigma_t-d_{ci}\right)\)

%\begin{displaymath}
%h_{ci}(t) = e^{\LARGE{\left({d^2_{ci}}/{2\sigma^2_t}\right)}}\cdot1\left(\sigma_t-d_{ci}\right)
%\end{displaymath}

%---------------------------------------Conclusions --------------------------------------------------------------------------

\section{Conclusions and Future Work}
In this article we classified the monte-carlo gamma ray data of the MAGIC experiment, using MLP and SOM. Both the networks shown good classification results.
 
The advantage of the SOM algorithm is that it needs no training vectors to find the groups in the data set {\em i.e.} it clusters the data set in an automatic way, but the disadvantage of this technique is that it cannot label the data groups found. At the other hand MLP based on supervised technique identifies the group labels, but the training session could be longer.

The proposal for the future work will be combining  MLP and SOM techniques. The combination of both techniques could yield better results. First train the data set with SOM, which yields in a clustered data set then use this data set to train the MLP to label the groups. This will significantly decrease the training period for MLP and thus makes the network to perform faster. 

%-------------------------------------bibilography--------------------------------------------------------------------

\end{document}